\documentclass[conference]{IEEEtran}
\IEEEoverridecommandlockouts

\usepackage[utf8]{inputenc}

\usepackage[backend=biber,style=ieee]{biblatex}
\usepackage{amsmath,amssymb,amsfonts}
\usepackage{graphicx}
\usepackage[hidelinks]{hyperref}
\usepackage{subfig}
\usepackage{xcolor}
\usepackage{listings}
\usepackage{verbatim}
\usepackage{graphicx}
\usepackage{multirow}
\usepackage{makecell}
\newcommand{\code}[1]{\textsc{#1}}

\newcommand{\chimera}{{\sc Chimera}}

\makeatletter
\newcommand{\linebreakand}{%
  \end{@IEEEauthorhalign}
  \hfill\mbox{}\par
  \mbox{}\hfill\begin{@IEEEauthorhalign}
}
\makeatother

\begin{document}

\title{Towards the Development of Entropy-Based Anomaly Detection in an 
Astrophysics Simulation}

\author{
\IEEEauthorblockN{Drew Schmidt}
\IEEEauthorblockA{\textit{Oak Ridge National Laboratory} \\
Oak Ridge, Tennessee\\
schmidtda@ornl.gov}
\and
\IEEEauthorblockN{Bronson Messer}
\IEEEauthorblockA{\textit{Oak Ridge National Laboratory} \\
Oak Ridge, Tennessee\\
bronson@ornl.gov}
\linebreakand
\IEEEauthorblockN{M. Todd Young}
\IEEEauthorblockA{\textit{Oak Ridge National Laboratory} \\
Oak Ridge, Tennessee\\
youngmt1@ornl.gov}
\and
\IEEEauthorblockN{Michael Matheson}
\IEEEauthorblockA{\textit{Oak Ridge National Laboratory} \\
Oak Ridge, Tennessee\\
mathesonma@ornl.gov}
}

\maketitle

\begin{abstract}
The use of AI and ML for scientific applications is currently a very exciting 
and dynamic field. Much of this excitement for HPC has focused on ML 
applications whose analysis and classification generate very large numbers of 
flops. Others seek to replace scientific simulations with data-driven surrogate 
models. But another important use case lies in the combination application of 
ML to improve simulation accuracy. To that end, we present an anomaly problem 
which arises from a core-collapse supernovae simulation. We discuss strategies 
and early successes in applying anomaly detection techniques from machine 
learning to this scientific simulation, as well as current challenges and 
future possibilities.
\end{abstract}

\section{Introduction}
\label{sec:intro}

Artificial Intelligence (AI) and Machine Learning (ML) have enjoyed a sustained 
explosion of popularity in recent years. Numerous applications from virtually 
every field have seen the benefits of AI for science~\cite{stevens2020ai}. Much 
of the excitement in the High Performance Computing (HPC) world has focused on 
large scale applications of AI methodology for analysis and classification of 
simulation data that consume gigantic amounts of floating point calculations. 
Some notable examples of this include~\cite{kurth2018exascale} and 
\cite{laanait2019exascale}. Others seek to accelerate scientific simulations by 
replacing part of them with pre-trained deep learning models. In this paper, we 
instead focus on an attempt at harnessing ML to improve the quality and 
accuracy of a scientific simulation. Specifically, we seek to build 
infrastructure which can detect anomalous events during the simulation run and 
alert the simulation that something has gone wrong.

Even in the most carefully architected software, mistakes occur. These
may be simple, easily understood bugs that are quickly patched, or they
may be deep sources of consternation that confuse teams for years. Sometimes 
errors occur that are not even the result of programmer mistakes; most people 
who have been around software engineers long enough eventually hear jokes about 
cosmic rays flipping bits in running programs. Of course, more than mere jokes, 
this can actually happen. It is far less likely to occur in systems with modern 
ECC RAM, although even then it is possible. Scientific simulations are no 
exception to these rules. They are usually quite large, very complicated, use 
difficult programming languages, and push the limits of boutique hardware. 
This compounds any inherent difficulty in finding and fixing software bugs.

\begin{figure}[t]
  \centering
  \subfloat[A valid image.]{
    \includegraphics[width=.45\textwidth]{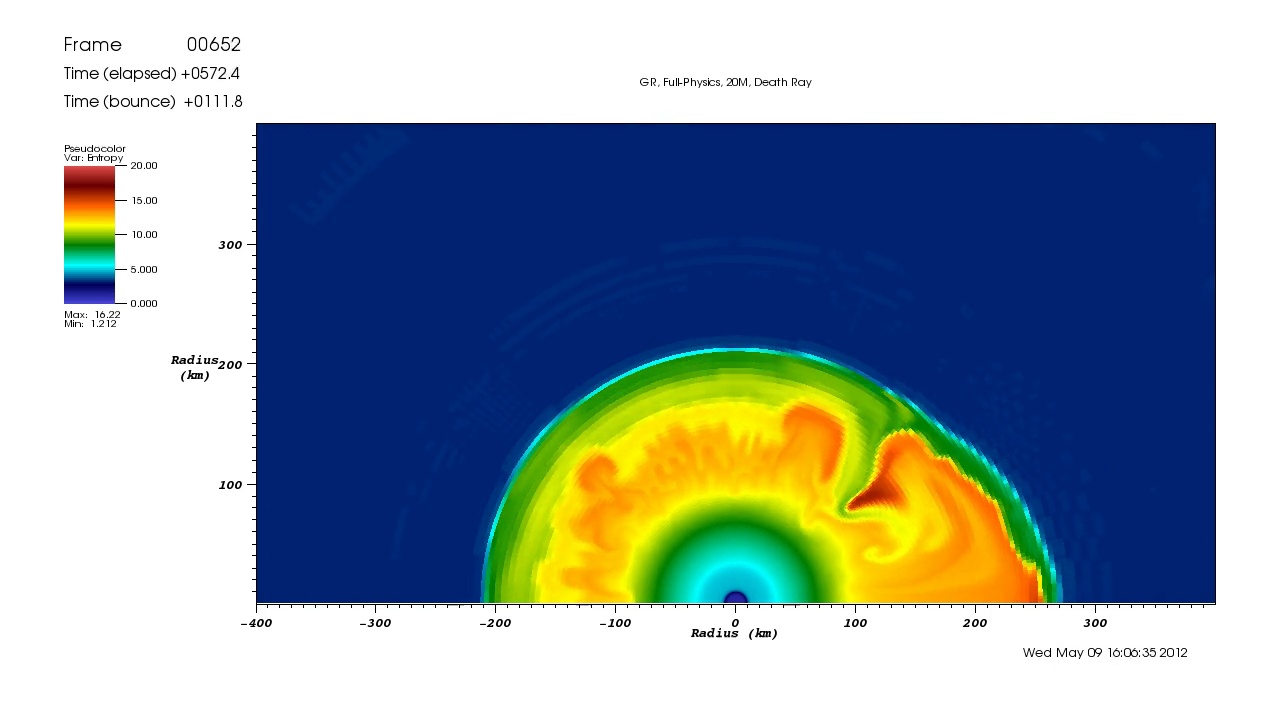}
    \label{fig:valid}
  }
  \hfill
  \subfloat[An anomalous image exhibiting the death ray.]{
    \includegraphics[width=.475\textwidth]{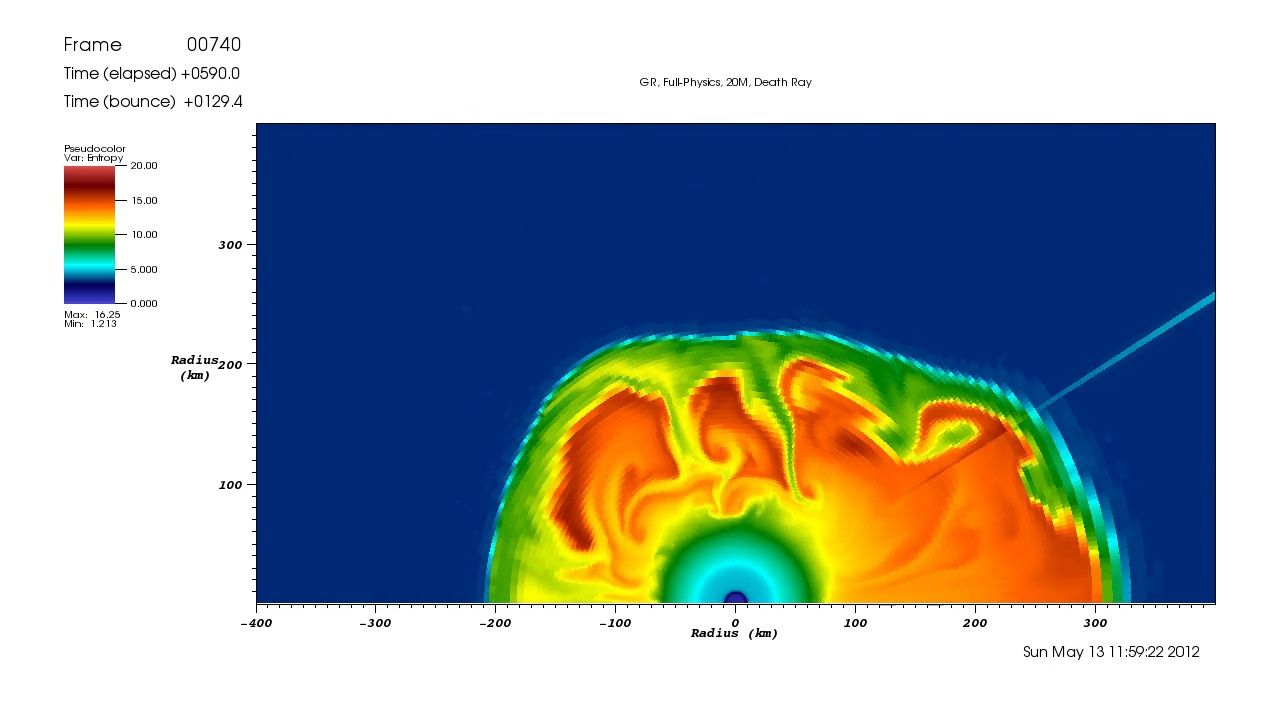}
    \label{fig:deathray}
  }
  \caption{Example images from the simulation. One is a valid image, while the 
other exhibits the death ray phenomenon.}
  \label{fig:example}
\end{figure}

We will be examining the outputs of the simulation code 
\chimera~\cite{messer2007petascale}. We will discuss the code and its 
scientific motivation in more depth in Section~\ref{subsec:chimera}. For the 
moment, we motivate only the anomaly problem. Figure~\ref{fig:example} shows 
two images from this program. These are from a simulation of the collapse and 
explosion of a massive star in a core-collapse supernova explosion 
\cite{Bruennetal2020}. Figure~\ref{fig:valid} shows an example of a typical 
time 
step from a running simulation, while Figure~\ref{fig:deathray} shows an 
example of an anomaly. The anomaly is the small line shooting off the edge of 
the convective interior of the star at roughly 45 degrees from the zenith is 
jokingly referred to as a ``death ray''. It was clear on inspection the death 
ray was not physical and could not have been the result of any reasonable 
evolution to that point.

Ideally, the simulation would be able to detect one of these anomalies as it 
occurs, and then either exit the simulation or perhaps rewind a few time steps 
to try again. And while it is easy for the human eye to see that something has 
gone wrong, detecting this from inside the simulation is very challenging. To 
try to solve this problem, we will turn to techniques from computer
vision and data analysis. The goal is to produce a model or framework which can 
be called by future runs of the simulation as a black box oracle to determine 
whether or not something has gone wrong. We tackle this with several 
approaches, each with different strengths and weaknesses.

One aspect of particular interest, the technical aspects of which we explore in
Section~\ref{sec:combining}, is the value of the simulation recovering from the 
detection of an anomaly. The simulation is, in principle, capable of rewinding 
by reverting to a recent checkpoint. This is in contrast to physical 
experiments which are often terminated in an unrecoverable state, destroying 
the experiment. Here, the experiment continues.

Although this work is still early and not fully finalized in a production 
environment, we have learned some valuable lessons which we believe are of broad
interest to anyone working on similar problems of rare event detection for 
scientific simulation. But before we can begin in earnest, we first discuss the 
scientific problem at length.
\section{Background}
\label{sec:background}

\subsection{The Scientific Problem}
\label{subsec:chimera}
Core-collapse supernovae are among the most energetic events in the Universe, 
releasing 10$^{53}$ erg of energy on timescales of a few tens of seconds. They 
produce and disseminate many of the elements heavier than helium, making life 
as 
we know it possible. They mark the birth of neutron stars and black holes.  

After several million years of evolution and nuclear energy release, a massive 
star's core is composed of iron (and similar `iron-peak' elements) from which 
no 
further nuclear energy can be released. Above this iron core remains a shell of 
silicon (Si). At the base of the Si-shell, nuclear burning continues, growing 
the iron core below. When the mass of the iron core reaches the limiting 
Chandrasekhar mass, it starts to collapse.

During the collapse, the inner core will become opaque to neutrinos and surpass 
the density of atomic nuclei ($\gtrsim$$2.5\times10^{14}\,g/cm^{3}$) reaching 
densities where individual nuclei merge together into nuclear matter. Above 
nuclear density, the nuclear equation of state (EoS) stiffens and the core 
rebounds like an over-compressed spring, launching a bounce shock from the 
newly 
formed neutron star (a proto-NS; PNS). The shock progresses outward through the 
rest of the infalling core, heating and dissociating the infalling nuclei to 
free nucleons and radiating a large burst of neutrinos. Thermal energy removed 
from the shocked material by neutrinos and nuclear dissociation halts the 
progress of the shock rendering it a standing accretion shock with a radius of 
100--200~km about 50~ms after it is launched. In this accreting state, the 
inner regions of the star continue to collapse and pass through the shock. The 
shocked, still infalling matter is dissociated and much of it settles on the 
PNS. Heating due to accretion onto the PNS drives the emission of neutrinos of 
all three flavors. Matter heating from neutrino interactions then produces 
large-scale convective overturn in the shock-heated mantle below the shock. 
Convection directly beneath the shock serves to markedly alter the state of the 
matter undergoing reheating via neutrino energy deposition 
\cite{HeBeHi94,BuHaFr95,JaMu96,FrWa04,BuRaJa06,BrDiMe06} relative to the 
spherically symmetric case. This inherently multidimensional effect allows 
simultaneous downflows that fuel the neutrino luminosities by accretion and 
upflows that bring energy to the shock. 

\chimera\ is a multi-dimensional radiation hydrodynamics code designed to study 
core-collapse supernovae. The name \chimera\ originates in its combination of 
three, separate, mature codes.  The primary code modules evolve the stellar 
fluid dynamics (MVH3), the ``ray-by-ray''  neutrino transport (MGFLD-TRANS), 
and 
the thermonuclear kinetics (XNet). These three ``heads'' are augmented by a 
sophisticated equation of state for nuclear matter. Hydrodynamics are evolved 
using a dimensionally split, piecewise parabolic method (PPM) --- a version of 
the publicly available astrophysics PPMLR hydrodynamics code VH1. Self-gravity 
is computed via multipole expansion in spherical harmonics. Spherical symmetric 
 corrections to gravity for GR  replace the non-GR (Newtonian) monopole 
($\ell=0$) term.

In the lower density regions outside the Fe-core, physical conditions require 
the use of a nuclear network to evolve the time-dependent abundances of nuclei.
The nuclear network incorporated into \chimera\ is the publicly available 
nuclear network code XNet. XNet solves, for each non-equilibrium zone, a 
coupled 
system of non-linear ODEs (one for each nuclear species) for the time evolution 
of the nuclear abundances.

Transport of neutrinos is computed via multi-energy angular moments of the 
neutrino distribution function in a multi-group, flux-limited diffusion 
approximation (MGFLD). The neutrino--matter interactions are a modern set that 
include scattering on electrons and free nucleons with energy exchange, 
emission 
and absorption on free nucleons and an ensemble of nuclei in NSE, and 
neutrino--anti-neutrino pair emission from nucleon-nucleon bremsstrahlung and 
$e^+e^-$-annihilation.

\chimera\ has been used on large supercomputer platforms for the past 15 years. 
Typical runtimes to complete a simulation starting from initial core collapse 
to 
3 seconds of physical time post-bounce have only become possible in 3D over the 
past few years. Even then, a typical \chimera\ simulation requires hundreds to 
thousands of compute cores for wallclock times of months (including queueing 
times and true runtime) to evolve the first seconds of an explosion. Because of 
this expense, close attention must be paid to each simulation during the entire 
run, as any errors that do not halt execution can lead to significant waste of 
resources. 
\subsection{Mapping The Simulation to Images}
The \chimera code incorporates detailed three dimensional hydrodynamics, a 
neutron reaction network, neutrino transport, and other physics that are 
necessary to model core-collapse supernovae. The complex interactions of the 
various components has made it difficult to determine the root cause of the 
death ray creation. However, it is possible to restart a simulation from a 
previously saved timestep and continue execution without encountering a 
death ray. This suggests that it is possible that a double-bit or multi-bit 
error has been encountered on the system and may be the cause of the anomaly.

In order to detect the anomaly, we use images created by the VisIt visualization 
package~\cite{childs2012visit}. The simulation domain is sliced to create a 2D 
image from the larger domain. The plot creation process naturally handles the 
mapping from a native polar coordinate simulation with varying grid resolutions 
to a rectangular image. The entropy variable of the simulation is used and 
linearly mapped to color by fixing the range by setting a minimum and maximum 
value for every time step. By fixing the viewport, the slice will be created in 
a common simulation space. This ensures that consistent images are used for the 
anomaly detection training and in subsequent simulations for the actual 
detection for a death ray.

This may seem a bit complicated; and perhaps on some level it is. But recall 
the key issue illustrated in Figure~\ref{fig:example}. The problem is easily 
detected by the human eye, even by someone uninitiated in the daunting field of 
astrophysics. Yet internally to the simulation, this is an exceedingly 
difficult problem to detect and correct. So our goal is to develop an AI 
strategy which can do what the human eye can do. We elucidate this strategy in 
the following section.
\section{Methodology And Results}
\label{sec:methodology}
\subsection{Preprocessing}
Because our dataset consists of images, a great deal of care and
pre-processing specific to image problems must be addressed. However,
the core principles outlined here apply beyond image problems.

The images are renderings of the matter entropy, a commonly used diagnostic 
in supernova simulations. The use of entropy to describe the matter allows 
clear delineation of the hot, shocked matter above the well-ordered (but, still 
quite hot) PNS. Our dataset consists of 813 total images, 81 of which are 
anomalous. This leaves 732 valid images.

Because our data are RGB images, we of course apply the standard
pre-processing before transforming them into multi-dimensional arrays.
Also, not all of the image is necessary, or perhaps even helpful for
training. As can be seen in Figure~\ref{fig:example}, there is a great
deal of extra information, such as the plot legend and timestamps, as
well as lots of white space. For our purposes, we restrict our analysis
to only the pixels contained in the intervals $[122, 601]$ for the
$x$ coordinates, and $[257, 1212]$ for the $y$.

After doing standard image pre-processing, our next goal is to extract
useful features from the images. To do this, we will use a convolutional
neural network (CNN), which are notoriously useful for computer vision.
Specifically, we use a ResNet-50~\cite{he2016deep} pre-trained on
ImageNet~\cite{deng2009imagenet} as our CNN. To borrow the overly fanciful and 
evocative language of the NN practitioners, we are essentially borrowing the 
object recognition capabilities that the network already learned and using them 
to recognize the important bits of the images in our dataset.

Next is the matter of developing a classifier. This is complicated enough to 
warrant a careful evaluation, which is the topic of the following subsection. 
We will use the extracted image features from the pre-trained ResNet-50 
with tabular classifiers. One may wonder why we do not use a neural network, 
such as an autoencoder~\cite{kramer1991nonlinear} here. In fact, we tried 
several such strategies, but with dismally poor results. Therefore, the final 
pre-processing amounts to reshaping the multi-dimensional array to a
matrix of abstract features (even more so than those obtained by the
neural network). The matrix will consist of one row per image, giving a
dimension of $732\times 921600$.
\subsection{Model Development Strategies}
There are many approaches to model development we could employ. One approach 
would be to use supervised learning methods. However, this has several 
drawbacks. For more general anomaly detection strategies, it potentially biases 
the classifier to particular types of anomalies present only during training 
because it requires specific labeling of the data.

More specifically to the problem at hand, we essentially only have one anomalous 
event. Recall, the data consists of 732 valid images, followed by the formation 
of the death ray. Each image starting from 733 onward is a continuation of the 
single anomalous event recorded in the data. The death ray expands a bit, 
experiences some of the same kinds of development as the supernova itself, but 
ultimately it is a single point of failure. The later images are incorrect, but 
they are not the demonstrations of failure we seek to discover. This makes 
models which were developed specifically rare event detection more attractive 
than general supervised learning classifiers.

Therefore, for our model development, we will employ unsupervised approaches. 
All of our experiments use one-class SVM~\cite{chen2001one} and isolation 
forests~\cite{liu2008isolation}. Using SVM is attractive because of its 
simplicity and its intuitively geometric behavior. The isolation forest is a 
powerful technique that is conceptually similar to the well-known random forest 
model in that it too is created from an ensemble of weaker binary classifiers. 
And like its cousin the random forest, it is known to perform very well for its 
intended purpose.

Finally, we will use these in two distinct ways: an offline approach and an 
online approach. For the offline approach, we will train on all the data up 
front as a post hoc process. For the online approach, we will pretend that the 
full dataset is not actually available to us at training time, as though we 
were working with the simulation in situ. The development of these strategies 
is provided in the subsections to follow.
\subsection{Offline Approaches}
Because our feature space is so large, we begin by applying principal 
components analysis (PCA)~\cite{wold1987principal} to position the data into an
optimally loaded subspace for developing our classification model. However, 
since we are only training on the non-anomalous data, we do not need to concern 
ourselves with the usual issues in principal components 
regression~\cite{sutter1992principal}, or consider advanced techniques like 
sufficient dimension reduction~\cite{cook2009likelihood}.

\begin{figure}[t]
  \centering
  \includegraphics[width=.475\textwidth]{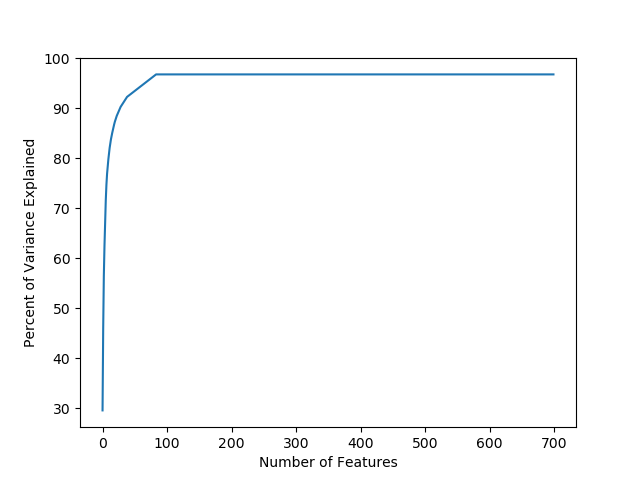}
  \caption{The percentage of variance explained plotted against the number of 
PCA rotated features.}
  \label{fig:scree}
\end{figure}

After applying the typical mean and standard deviation scaling column-by-column 
(transforming the features into z-scores), we perform PCA. A typical scree plot 
analysis on a plot such as that in Figure~\ref{fig:scree} would suggest 
retaining somewhere on the order of 100 features. However, experimentation with 
the number of features showed that we achieve uniformly better classifier
results as we increase the number of features, up until around 500. For
our final model, we retain 512 features, transforming the dataset into a
$732\times 512$ matrix. For testing, we use a pool of 40 randomly selected 
valid images and 20 invalid images.

As noted in the preceding subsection, the models we train on the images are 
one-class SVM and isolation forest. For each, we optimized the hyperparameters 
of using Bayesian optimization~\cite{snoek2012bayesopt}, minimizing the models' 
misclassification rates. For the one-class SVM, we use a radial basis kernel 
function with coefficients $\gamma = 0.001$ and $\nu = 0.01$. For the isolation 
forest, we used 125 base estimators for the ensemble, $contamination = 0.0$, 
and $max\_sample = 1.0$. After optimizing the model hyperparameters, the SVM 
achieves $95\%$ accuracy on the test set and the isolation forest gets $100\%$ 
accuracy on the test set. The confusion matrix for the SVM's predictions on the 
test set is given by Table~\ref{tab:svm_confusion}. The confusion matrix for 
the isolation forest's test set predictions is given by 
Table~\ref{tab:isof_confusion}.

\begin{table}[ht]
\centering
  \begin{tabular}{cc|cc}
    \multicolumn{1}{c}{} &\multicolumn{1}{c}{} &\multicolumn{2}{c}{Test Set 
Anomalous} \\
    \multicolumn{1}{c}{} & 
    \multicolumn{1}{c|}{} & 
    \multicolumn{1}{c}{Yes} & 
    \multicolumn{1}{c}{No} \\ \hline
    \multirow[c]{2}{*}{Predicted}
    & Yes & 37 & 3  \\[1.25ex]
    & No  & 0  & 20 \\ \hline
    \end{tabular}
  \caption{SVM Confusion Matrix}
  \label{tab:svm_confusion}
\end{table}

\begin{table}[ht]
\centering
  \begin{tabular}{cc|cc}
    \multicolumn{1}{c}{} &\multicolumn{1}{c}{} &\multicolumn{2}{c}{Test Set 
Anomalous} \\
    \multicolumn{1}{c}{} & 
    \multicolumn{1}{c|}{} & 
    \multicolumn{1}{c}{Yes} & 
    \multicolumn{1}{c}{No} \\ \hline
    \multirow[c]{2}{*}{Predicted}
    & Yes & 40 & 0  \\[1.25ex]
    & No  & 0  & 20 \\ \hline
    \end{tabular}
  \caption{Isolation Forest Confusion Matrix}
  \label{tab:isof_confusion}
\end{table}

\begin{table}[t]
  \centering
  \begin{tabular}{rrrr}
  \hline
    Model & Normal & Anomalous & Overall\\\hline
    SVM & 662 & 81 & 91.4\% \\
    Isolation Forest & 732 & 81 & 100\% \\\hline
  \end{tabular}
  \caption{Classifier accuracy. Numbers shown in the ``Normal'' and 
``Anomalous'' columns are counts, i.e. the number of correctly classified 
images of each type. The ``Overall'' figure is the weighted classification 
accuracy.}
  \label{tab:accuracy}
\end{table}

The results of the model performance are summarized in Table~\ref{tab:accuracy}. 
Both models classify all anomalous images correctly. However, the SVM struggles 
with the normal, non-anomalous data. The isolation forest on the other hand 
correctly predicts both normal and anomalous data. Because the one-class SVM 
struggles significantly more compared to the isolation forest, for now we 
proceed only with the latter.

Next, we run a series of experiments to test the capacity the trained model has 
for transfer learning. We begin with using standard image analysis tricks, like 
mirroring the images. Testing horizontal flips alone, we achieve an overall 
classifier accuracy of 251/732 on the normal images, and 81/81 of the 
anomalous. Testing vertical flips (which are impossible for the simulation to 
actually produce), the model achieves 0/732 on the normal images and 81/81 on 
the anomalous. Finally, for images which are both vertical and horizontally 
flipped, the model again scores 0/732 normal images and 81/81 anomalous.

Originally we intended to expand the size of the training set by generating 
false death rays on existing non-anomalous images. However, this strategy seems 
to suggest that there is little reason to proceed. Essentially, the model 
is radically overtraining to the data so that it greatly struggles to detect
anything but what it has seen. This is why it is so ``good'' at detecting 
anomalies, and almost certainly means that we can not use this trained model on 
future simulation runs expecting good predictions on the new data. All of this 
suggests that a different strategy is warranted.
\subsection{Online Approaches}
Due to the limitations of the offline approaches in achieving transfer, we 
proceed with an online approach. Since we actually already have the images, we 
will pretend that new images are being generated, and develop a new model each 
time. So it is not truly online, but the strategy could easily be adapted to an 
actual online environment. It is important to note that we are not, strictly 
speaking, trying to develop a \emph{model} as the primary outcome, but instead a 
pool of images which we believe are valid. At each step, the model is discarded 
after making a prediction.

As a baseline, we will start with the model development strategy used in the 
offline setting, training a new model each time a new image is made available. 
At that time, we take as a matter of faith that all prior images are valid, 
train a model like before, and then make a prediction on the new image. 
Unfortunately, this does not work even a little bit. The models are never able 
to correctly identify a next image when the image is valid, similar to the 
problem experienced in the offline setting.

\begin{figure}[ht]
  \centering
  \subfloat[All model predictions.]{
    \includegraphics[width=.45\textwidth]{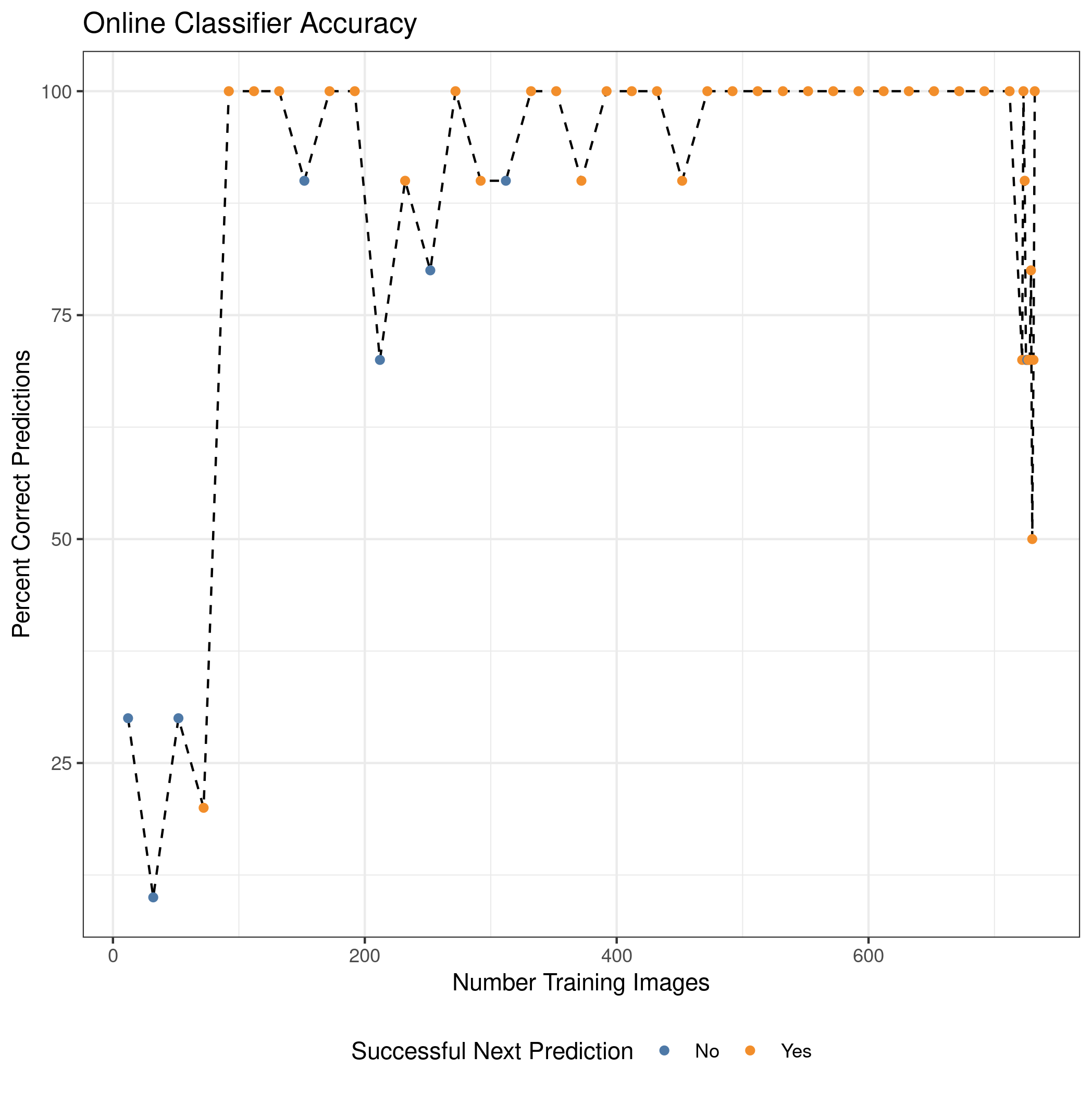}
    \label{fig:online_preds_all}
  }
  \hfill
  \subfloat[Model predictions near and including anomalous images.]{
    \includegraphics[width=.45\textwidth]{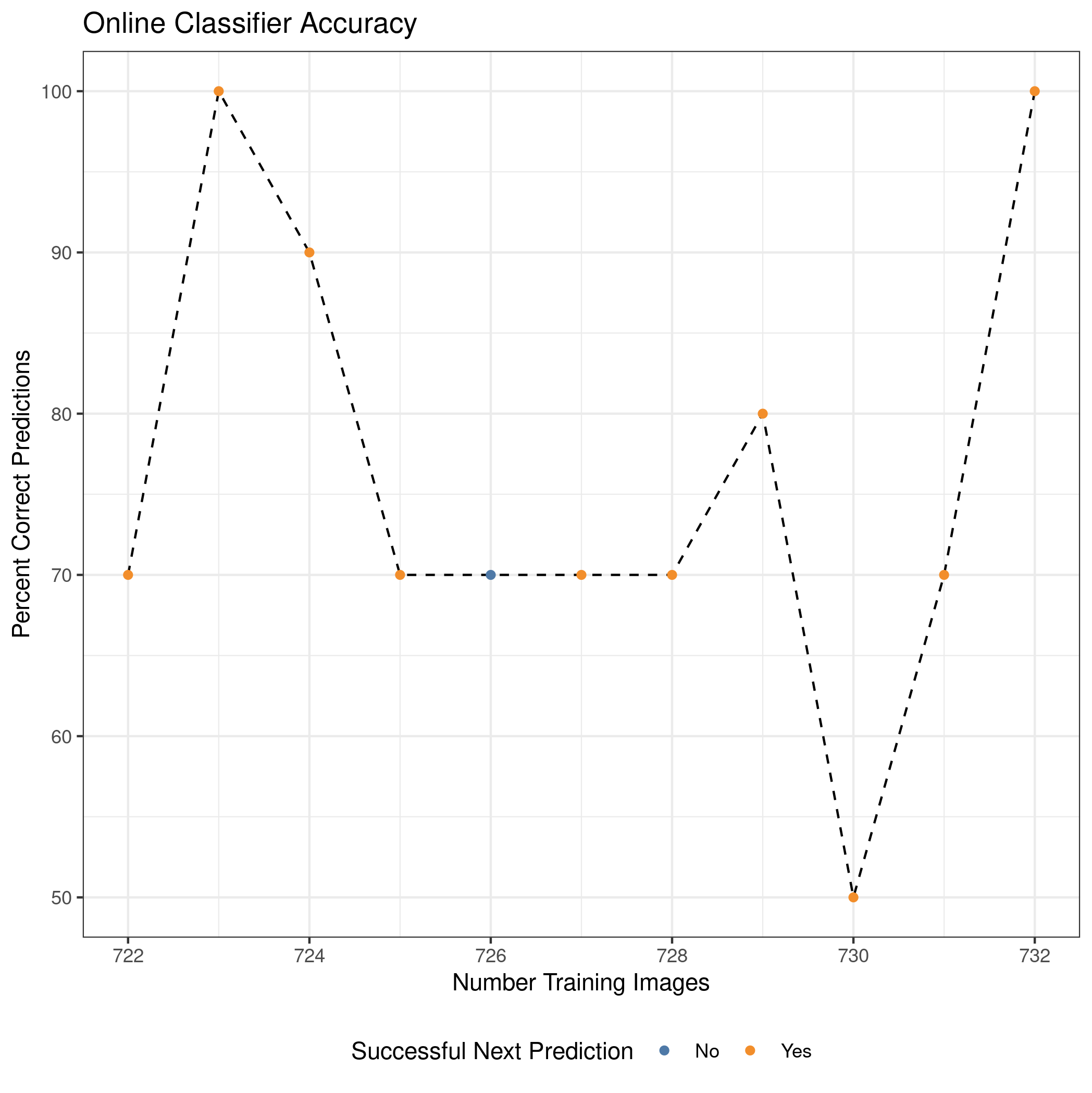}
    \label{fig:online_preds_big}
  }
  \caption{Percent of correct predictions for the series of models from the 
online training. Each model makes a prediction on the last image it saw plus 
nine images it has not seen. Models which correctly classify the first unseen 
image are colored orange, and those which fail are colored blue. The line is 
dashed for better visibility.}
  \label{fig:online_preds}
\end{figure}

So next, we try developing the models without performing the PCA-based 
dimension reduction. This radically increases the training time as the dataset 
grows, which is a serious issue for the true online setting. Nevertheless, 
proceeding with this shows some promise. Figure~\ref{fig:online_preds} shows 
the results of proceeding in this way. For each model, we test on the last 
image that the model saw, and find that every time it correctly predicts that 
image as valid. We then make predictions on the next 9 images, each of which 
was unseen by the model. This is helpful to ensure that the model is not simply 
throwing away everything it has not yet seen. However, the only real prediction 
that matters is for the first image it has not yet seen. In the plot, we color 
those which are able to make that first true prediction correctly orange, or 
otherwise blue.

Early on, the classifiers struggle a great deal to make good predictions at 
all, as can clearly be seen in Figure~\ref{fig:online_preds_all}. But by the 
time we reach 100 images in the training set, the models score well on their 
test sets overall. However, we still see many failures on the prediction of the 
first unseen image until roughly 300 images constitute the training set. After 
that, they largely disappear. For the models which predict on the anomalous 
images, shown in clearer detail in Figure~\ref{fig:online_preds_big}, overall 
we see good performance. Except in one case, specifically the model trained on 
725 images, they are all able to correctly predict the first unseen image. This 
includes the first anomalous image, correctly detected by the model trained on 
732 images. And although there is no simple way to demonstrate this in the 
plot, half of the models were able to correctly detect the first anomalous 
image in their test set of 9 unseen images. Why the model trained on 725 images 
fails is not at all clear to us.

As expected, there is a training data accumulation problem with this approach. 
As we have seen, the initial few models are of poor quality, since they have 
seen only a handful of snapshots of the simulation. So for spotting early death 
rays, this is not a good strategy, since we essentially have to build into the 
decision-making system a distrust of the early models. The best way to overcome 
this shortcoming is not immediately obvious.

Finally, a few comments about this strategy compared to the offline one above. 
First, we experimented with SVM models as before; but also as before, they were 
seriously outperformed by the isolation forest. In addition, we found no major 
improvement to the transfer problem by dropping the PCA step in the offline 
setting. Although it improves somewhat, we find that the online variant shows 
the most promise. However, the training time is radically more expensive than 
before, taking hours to fit a single model. And there are not to our knowledge 
any alternative, high-performance isolation forest packages available which 
could credibly diminish the computing time to a sufficiently short interval.
\section{Invoking the Anomaly Detector from the Simulation}
\label{sec:combining}

Here, we discuss some strategies for calling the ML framework from the 
simulation as it produces images. We must do so in such a way that it can inform 
the simulation if something may have gone wrong. There are several ways in which 
we could do this, and weighing a few of their pros and cons is warranted.

The simplest option is to simply spawn a new ML process every time a new image 
is generated. That is, each time the simulation (a C, C++, or Fortran code) 
produces a new image in need of classification, it will simply spawn a new 
classifier process via a call to \texttt{system()} (with appropriate wrappers in 
the case of Fortran). The simulation, which may well be a large, distributed 
code communicating via MPI~\cite{walker1996mpi}, need only call this on
one of its processes. The simulation must wait on hold for the result. But 
after it is returned to the calling process, that process can then broadcast 
the result to the remaining processes and act accordingly.

This approach makes the most sense if the run time for the ML process is very 
short. This aligns with our offline model, where although the run time cost to 
pre-process many images and perform training is non-trivial, classifying a 
single image is quite quick. On an ordinary workstation, the total run time to 
classify a new image is about 5 seconds, including the cost of starting up a new 
interpreter and loading the necessary libraries and stored data/models. 

Our model codes anomalous images as $-1$ and acceptable ones as $1$. So we 
merely need to have the interpreter exit with code $0$ if the new image is not 
detected as anomalous (the model prediction is $1$), and some non-zero error 
code otherwise. Putting this into practice, the C code for the sim to call might 
look as simple as:
\begin{lstlisting}[language=C]
int classify(char *img_path){
  char cmd[256];
  sprintf(cmd, "/bin/interpreter \
    classify_img.script %
  return system(cmd);
}
\end{lstlisting}
As noted before, this wrapper is easily made Fortran-compatible via the usual 
\texttt{ISO\_C\_BINDING}. For this then, the simulation workflow might look 
something like:
\begin{enumerate}
\item Simulate time step.
\item Generate image, call \texttt{classify()} as outline above.
\item If the return value is 0, continue. Otherwise, rewind to the last
  checkpoint and try again.
\end{enumerate}
If the $\approx 5$ second run time for classifying a new image in this set is 
intolerable, note that roughly half of it is devoted to starting a new 
interpreter and loading the requisite libraries, data, and models. It is 
possible to create a persistent process that could be polled by the simulation 
with, say, a ZeroMQ~\cite{hintjens2013zeromq} REQ/REP socket between the 
calling process of the simulation and the analysis process. The simulation 
would essentially tell the persistent anomaly detector process that it has a new 
image at a specified location (communicated over the socket), and the anomaly 
detector would then evaluate and reply on the same socket.

Neither of these strategies is viable for the online variant due to its 
radically longer training times. In this case, it would be more reasonable to 
set aside some compute resources on the supercomputer. That is, some subest of 
the nodes could be carved out specifically for the processing and training of 
images. This training can work independently of the sim in an overlapped way. 
For the first image, the analysis processes would wait. After it is generated, 
the simulation could proceed to the next time step while the analysis process 
operates on the first image. Then the simulation generates the second image, 
proceeds to the third time step, all while the analysis process operates on the 
second image. And on and on.

Launching this kind of dual application with MPI is straightforward in 
principle. Say $N_1$ resources were devoted to the simulation code, while $N_2$ 
were devoted to the classifier script. Then one might do something like:
\begin{lstlisting}[language=C]
mpirun -np ${N1} sim : -np ${N2} /bin/interpreter classify_img.script 
/sim/img/path
\end{lstlisting}
In this case, the simulation and the anomaly detector processes each maintain 
two MPI communicators. One is the main communicator, say 
\texttt{MPI\_COMM\_WORLD}, and the other is a ``local'' work communicator only 
for processes of the same type. This latter communicator is easily constructed 
via \code{MPI\_Comm\_split}, where say, the simulation uses color 0 and the 
anomaly detector uses color 1. In practice this is complicated somewhat by the 
boutique MPI launcher replacements found on many supercomputers. But the 
principle remains.

Ideally the analysis process would finish evaluating new images far faster than 
the simulation of each time step. In this case, deciding when to coordinate the 
results of the analysis process with the simulation is moot. If the anomaly 
detection is far slower, then this creates a bottleneck from the point of view 
of the simulation. That is likely the case for our particular model-building in 
the online case. Worse, the isolation forest model fitter we use is not 
scalable even within a single node. This creates the largest hurdle to an 
actual application of online strategy.
\section{Conclusions and Future Work}
\label{sec:conclusions}

In this paper, we outlined multiple strategies for developing an entropy-based 
anomaly detector for a core-collapse supernovae simulation. While the 
discussion was focused around our particular scientific dataset, these 
principles could easily be applied to other scientific domains. We provided 
two approaches, outlined the advantages and the shortcomings of each, and then 
closed with strategies for integrating them with the simulation.

Neither approach was fully satisfactory, although for greatly different 
reasons. We think it may be possible to salvage the offline approach by 
including more data in the training set. This could be done by generating more 
images from additional simulation runs, for example. However, in some ways this 
seems self-defeating. It may be possible to sufficiently expand the dataset 
using more image transformations. For the online approach, the biggest obstacle 
to moving this strategy into production is the training time. If this approach 
is the only viable path forward, then this would necessitate the development of 
a high-performance isolation forest model fitter, which is a difficult task unto 
itself.

In a truly online environment, that approach would have access to more 
information. Namely, we could use the raw data from the simulation as opposed 
to an image snapshot. It is not guaranteed that this will improve the model 
accuracy, but it would certainly complicate things regardless.
\section*{Acknowledgements}
This research used resources of the Oak Ridge Leadership Computing Facility at 
the Oak Ridge National Laboratory, which is supported by the Office of Science 
of the U.S. Department of Energy under Contract No. DE-AC05-00OR22725.

This manuscript has been authored by UT-Battelle, LLC, under contract 
DE-AC05-00OR22725 with the US Department of Energy (DOE). The US government 
retains and the publisher, by accepting the article for publication, 
acknowledges that the US government retains a nonexclusive, paid-up, 
irrevocable, worldwide license to publish or reproduce the published form of 
this manuscript, or allow others to do so, for US government purposes. DOE will 
provide public access to these results of federally sponsored research in 
accordance with the DOE Public Access Plan 
(\url{http://energy.gov/downloads/doe-public-access-plan}).
\appendices

\section{Software Environment}
\label{sec:artifacts}

Our ML framework is written in the python programming 
language~\cite{rossum1995python}. Our package, including code and data,
is available at \url{https://code.ornl.gov/va8/autoencoder}.
We use python for easiest access to the various Deep Learning frameworks for
extracting ResNet features, namely
Tensorflow~\cite{abadi2016tensorflow}, keras~\cite{chollet2015keras},
and pytorch~\cite{paszke2019pytorch}. We also use keras, as well as 
Numpy~\cite{oliphant2006guide}, for much of the image pre-processing. 
We use Scikit-learn~\cite{pedregosa2011scikit} to apply PCA, as well as in
fitting the SVM and Isolation Forest models.
The Bayesian Optimization was performed by the
scikit-optimize package~\cite{head2018scikitopt}.
It is worth noting that none of the analysis in
Section~\ref{sec:methodology} relies on features native or exclusive
to any particular programming language or package. Finally, we note the package 
versions used during model development:

\begin{itemize}
    \item python 3.8.2
    \item keras 2.4.0
    \item numpy 1.17.4
    \item Scikit-learn 0.22.2.post1
    \item scikit-optimize0.8.1
    \item tensorflow 2.3.0
\end{itemize}

\printbibliography

\end{document}